\documentclass[10pt,twocolumn,letterpaper]{article}

\usepackage[pagenumbers]{cvpr} % To force page numbers, e.g. for an arXiv version

\usepackage{tabularx,fvextra}

\definecolor{cvprblue}{rgb}{0.21,0.49,0.74}
\usepackage[pagebackref,breaklinks,colorlinks,allcolors=cvprblue]{hyperref}
\usepackage{booktabs,makecell,threeparttable,adjustbox}
\usepackage{multirow}
\usepackage{tabularx}
\usepackage{fvextra}
\definecolor{darkgreen}{RGB}{0,128,0}

\def\confName{CVPR}
\def\confYear{2027}

\title{WINSO: A Large-Size Very-High-Resolution Remote Sensing Dataset for \\ Wind Turbine Detection and Solar Panel Segmentation}

\author{
Di Wang$^{1}$,
Ke Li$^{1}$, %\thanks{Corresponding authors},
Zhaokun Li$^{1}$,
Weiping Ni$^{1}$, %\footnotemark[1],
Junzheng Wu$^{2}$,
Yahu Liu$^{3}$,
Quan Wang$^{1}$,
Xinbo Gao$^{1}$\\
$^1$Xidian University \quad
$^2$Northwest Institute of Nuclear Technology \\
$^3$Northwestern Polytechnical University\\
}

\begin{document}
\twocolumn[{
\renewcommand\twocolumn[1][]{#1}% 
\maketitle 
\vspace{-6mm}
\begin{center} 
    \centering 
    \includegraphics[width=0.97\textwidth]{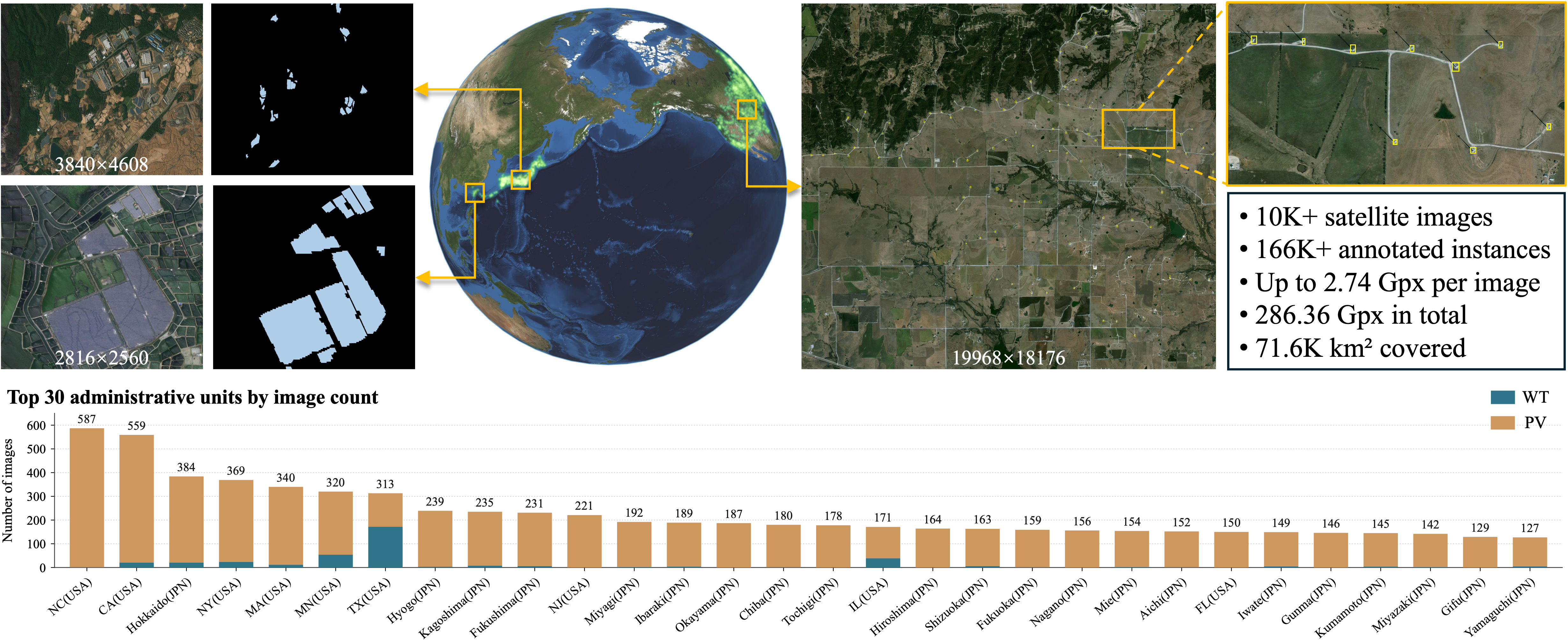} 
    \captionof{figure}{Overview of the proposed WINSO dataset.
    \textbf{Top:} Our dataset provides horizontal bounding boxes for wind turbines (WT) and segmentation masks for solar photovoltaic (PV) panels across the Asia-Pacific region.
    Gpx denotes $10^9$ pixels. The reported 71.6K~km$^2$ is the cumulative image area estimated at a GSD of 0.5~m.
    \textbf{Bottom:} Image counts for the top 30 administrative units, ranked by total image count. 
    Stacked bars distinguish WT and PV subsets, and numbers above the bars indicate total counts.}
    \label{fig:dataset_overview} 
\end{center}
}]
\maketitle

\begin{abstract}
Wind and solar energy are central to the global transition toward net-zero emissions and sustainable development.
Accurate spatial inventories of wind and solar facilities are essential for future energy planning.
Satellite remote sensing provides a practical and efficient means to acquire such geospatial data at scale.
However, reliable facility mapping from large-size satellite images remains constrained by the scarcity of carefully annotated datasets and the lack of systematic evaluation protocols.
To address these limitations, we introduce \textbf{WINSO}, a large-scale wind and solar benchmark based on very-high-resolution (VHR) satellite imagery with ground sampling distances (GSDs) of 0.5--0.7\,m.
WINSO comprises 10,962 large-size images sampled from diverse geographic locations across the Asia-Pacific region, covering at least 71,600\,$\text{km}^2$ and totaling 286.36 gigapixels, with individual image dimensions reaching 44,032$\times$62,208 pixels.
Through manual annotation and expert verification, a total of 166,473 instances are labeled using horizontal bounding boxes (HBBs) for wind turbines and polygon masks for solar panels.
To build baselines for clean-energy facility mapping, we evaluate state-of-the-art detection and segmentation models, specialized ultra-high-resolution (UHR) architectures, and vision-language models across different inference strategies. 
Furthermore, cross-dataset evaluations on two external public datasets illustrate the strong generalization capability of the WINSO.
WINSO establishes a standardized benchmark to advance automated clean-energy mapping, paving the way for wide-area monitoring of the clean-energy transition.
\end{abstract}    
\section{Introduction}
\label{sec:intro}
Renewable energy is a critical component of the global energy transition, with wind and solar facilities playing a central role in decarbonizing electricity generation~\cite{liu2023climate}.
They hold substantial significance in the domains of climate change mitigation, energy security, and sustainable development~\cite{lei2023co}.
As clean-energy installations proliferate rapidly, automated and accurate wind turbine detection and solar photovoltaic (PV) panel segmentation have become indispensable for supporting power grid planning, resource potential assessment, and ecological and environmental monitoring~\cite{kruitwagen2021global,paolo2024satellite}.

Benefiting from extensive spatial coverage, satellite remote sensing imagery has emerged as an important data source for mapping renewable energy facilities. 
However, as summarized in Tab.~\ref{tab:wind_solar_datasets}, existing benchmarks reveal practical \textbf{limitations in combining large instance volumes with fine-grained annotations}. 
General remote sensing datasets (\textit{e.g.}, DIOR~\cite{li2020object}, AI-TOD~\cite{wang2021tiny}) cover a broad range of object categories, with limited emphasis on renewable energy facilities.
While specialized benchmarks have emerged to address this gap, existing efforts in both wind and solar domains exhibit two primary compromises:
\textit{(1) For wind turbines}, recent large-scale benchmarks such as SatlasPretrain~\cite{bastani2023satlaspretrain} and TinyWT~\cite{zhu2024tinywt} achieve impressive instance counts, but rely on coarse-resolution imagery (\textit{e.g.}, 10\,m GSD) and point-level annotations. Notably, point labels alone do not encode turbine extent, limiting the geometric supervision available for rotor-size estimation and downstream tasks such as turbine capacity estimation and wake-effect modeling. 
\textit{(2) For solar panels}, high-resolution datasets such as DukePV~\cite{bradbury2016distributed} and BDAPPV~\cite{kasmi2023crowdsourced} predominantly rely on localized aerial photography, which exhibits non-trivial distribution shifts from orbital satellite imagery in target scale and viewing geometry. Meanwhile, larger-scale satellite collections such as GloSoFarID~\cite{yang2024glosofarid} often rely on model-inferred pseudo-labels, which can introduce errors in facility boundary delineation and area estimation.

Beyond annotation fidelity, existing benchmarks also face \textbf{limitations in preserving complete facility layouts and large-scale spatial context}. 
In real-world environments, renewable energy facilities often form spatial clusters, such as contiguous solar arrays and aligned wind turbines. 
However, as shown in Tab.~\ref{tab:wind_solar_datasets}, conventional datasets routinely rely on small, fixed-size image patches (typically 256--512 pixels) that fragment facility layouts and obscure broader geographic surroundings. 
Consequently, benchmarking directly on large-size imagery enables the evaluation of models under realistic, wide-area mapping scenarios, bridging the gap between patch-based evaluation and operational satellite monitoring.

To bridge these critical gaps, we construct \textbf{WINSO}, a specialized benchmark for wind turbine detection and solar panel segmentation in VHR satellite imagery.
As illustrated in Fig.~\ref{fig:dataset_overview}, WINSO comprises 10,962 large-size satellite images totaling 286.36 gigapixels (Gpx) and covering over 71.6K\,$\text{km}^2$.
Crucially, rather than relying on pre-cropped patches, WINSO preserves large-size satellite scenes, averaging 14,192$\times$11,484 pixels per image for wind farms and 3,517$\times$2,965 pixels for solar arrays, with individual images reaching up to 2.74\,Gpx. 
In total, WINSO provides 166,473 manually verified instances, consisting of 31,057 HBBs for wind turbines and 135,416 polygon masks for solar panels.
To ensure pristine annotation fidelity, all labels were produced by a team of 25 well-trained annotators and underwent rigorous, image-by-image visual validation by three remote sensing domain experts. 
Geographically, WINSO captures broad topographic and climatic diversity across three regions: the United States, Japan, and China's Taiwan region, encompassing varied real-world environments from mountainous ridges, desert expanses, and coastal areas to complex industrial parks and agricultural zones.

Equipped with WINSO, we benchmark general detectors, segmentors, specialized ultra-high-resolution (UHR) architectures, and vision-language models (VLMs) across diverse inference strategies. 
Our empirical findings reveal striking performance contrasts across these regimes: while local tiled inference achieves the highest accuracy at the expense of high latency, direct global inference suffers from a catastrophic performance collapse.
Encouragingly, specialized UHR architectures demonstrate strong promise by striking a favorable balance between throughput and spatial fidelity. 
Furthermore, we conduct cross-dataset generalization experiments to demonstrate that models trained on WINSO transfer effectively to external benchmarks and exhibit robust out-of-domain adaptability. 
We hope this benchmark establishes a standardized, reproducible testbed to advance the fundamental evaluation of clean-energy facility mapping in large-size satellite imagery.

In summary, our main contributions are as follows:
\begin{itemize}
    \item We introduce WINSO, a large-scale benchmark for wind turbine detection and solar panel segmentation in VHR satellite imagery. With 10,962 large-size images and 166,473 expert-verified instances across diverse geographic backgrounds, it offers a unique resource for wide-area renewable energy mapping.
    \item We benchmark general detectors and segmentors, specialized UHR methods, and VLMs under resizing and tiled inference regimes, uncovering critical speed-accuracy trade-offs and highlighting content-adaptive sparse processing as a promising direction for large-size remote sensing perception.
    \item We demonstrate WINSO's robust cross-dataset transferability for both wind turbine detection and solar panel segmentation, providing an open empirical foundation for automated renewable facility monitoring.
\end{itemize}

\begin{table*}[!t]
\centering
\caption{Comparison of WINSO with representative benchmarks.
$\dagger$ denotes the average image width and height.}
\label{tab:wind_solar_datasets}
\begin{threeparttable}
\begin{adjustbox}{max width=\textwidth}
\begin{tabular}{@{}lcccccccc@{}}
\toprule
\textbf{Dataset} &
\makecell{\textbf{WT}\\\textbf{labels}} &
\makecell{\textbf{PV}\\\textbf{labels}} &
\makecell{\textbf{Images}} &
\makecell{\textbf{Instances}} &
\makecell{\textbf{Annotation}} &
\makecell{\textbf{Image size} \\\textbf{(px)}} &
\makecell{\textbf{GSD} \\\textbf{(m)}} &
\textbf{Source} \\
\midrule
\multicolumn{8}{@{}l}{\textit{\textbf{General remote sensing benchmarks: energy-related categories only}}} \\[3pt]
DIOR~\cite{li2020object}                              & HBB   & -- & 1,616 & 5,363 & Manual & 800$\times$800 & 0.5--30 & Aer./Sat. \\
AI-TOD~\cite{wang2021tiny}                            & HBB   & -- & 55    & 533   & Reused & 800$\times$800 & --      & Aerial \\
SatlasPretrain~\cite{bastani2023satlaspretrain}       & Point & Polygon / Point & -- & WT: 354,120 \ PV: 58,216\tnote{a} & Hybrid & 512$\times$512 & 1, 10 & Aer./Sat. \\ 
STAR~\cite{li2025star}                                & OBB   & -- & 51 & 1,298 & Manual & 11,732$\times$9,788$^{\dagger}$ & 0.15--1 & Aerial \\
\midrule
\multicolumn{8}{@{}l}{\textit{\textbf{Dedicated wind turbine datasets}}} \\[3pt]
Overhead WT (NAIP)~\cite{kornfein2023closing}         & HBB            & -- & 1,200 & 1741             & --     & 608$\times$608              & 1  & Aerial \\
Multi-topography WT~\cite{hao_mi_2023_7808269}        & HBB\tnote{b}   & -- & 1,270 & 4,458            & Manual & --                & -- & -- \\
TinyWT~\cite{zhu2024tinywt}                           & Point          & -- & 8,871 & 736,426\tnote{c} & Hybrid & 513--6{,}731      & 10 & Satellite \\
\midrule
\multicolumn{8}{@{}l}{\textit{\textbf{Dedicated photovoltaic datasets}}} \\[3pt]
Duke California PV~\cite{bradbury2016distributed} & --    & Polygon & 601     & 19,433\tnote{d}  & Manual      & 5,000$\times$5,000 & 0.3         & Aerial \\
Multi-resolution PV~\cite{jiang2021multi}         & --    & Mask    & 3,716   & --               & Manual      & 256$\times$256     & 0.1/0.3/0.8 & Aer./Sat. \\
BDAPPV~\cite{kasmi2023crowdsourced}               & --    & Mask    & 46,132  & 20,989           & Manual      & 400$\times$400     & 0.1/0.2     & Aerial \\
H-RPVS~\cite{wang2023rooftop}                     & --    & Mask    & 5,866   & 16,925           & Manual      & 256$\times$256     & 0.15        & Aerial \\
PVP-Dataset~\cite{wang2023pvnet}                  & --    & Mask    & 4,640   & --               & --          & 512$\times$512     & --          & Satellite \\
AIR-PV~\cite{yan2023air}                          & --    & Mask    & 5,000   & --               & --          & 512$\times$512     & 0.8         & Satellite \\
GloSoFarID~\cite{yang2024glosofarid}              & --    & Mask    & 13,703  & --               & Hybrid      & 256$\times$256     & 10          & Satellite \\
GM-SEUS~\cite{stid2025harmonized}                 & --    & Mask    & 17,500  & 4,452            & Hybrid      & 256$\times$256     & 0.3--0.6    & Aerial \\
Maxar Germany PV~\cite{clark2023solar}            & --    & HBB     & 2,542   & 2,542            & Manual      & 416$\times$416     & 0.31        & Satellite \\
\midrule
\textbf{WINSO (Ours)} &
\textbf{HBB} & 
\textbf{Mask} &
\makecell{\textbf{10,962} \\ \textbf{(WT: 1,045 \ PV: 9,917)}} &
\makecell{\textbf{166,473} \\ \textbf{(WT: 31,057 \ PV: 135,416)}} &
\textbf{Manual} & 
\makecell{\textbf{WT: 14,192 $\times$ 11,484} \\ \textbf{PV: 3,517 $\times$ 2,965}}$^{\dagger}$ &
\textbf{0.5--0.7} &
\textbf{Satellite} \\
\bottomrule
\end{tabular}
\end{adjustbox}
\begin{tablenotes}[flushleft]
\scriptsize
\item[a] SatlasPretrain reports 354,120 WT point labels. Its PV count combines 32,803 solar-farm polygons and 25,413 rooftop-solar point labels.
\item[b] Multi-topography WT boxes jointly enclose the turbine body and its shadow.
\item[c] TinyWT point counts include repeated temporal observations and do not represent unique physical turbines. 
\item[d] Duke California PV uses the corrected v4 (2020) count of array polygons.
\end{tablenotes}
\end{threeparttable}
\end{table*}
\section{Related Work}
\label{sec:related_work}

\noindent \textbf{General Remote Sensing Datasets.}
Remote sensing benchmarks have steadily evolved to encompass increasingly diverse categories and realistic spatial extents~\cite{xuan2026dynamicvl,wang2026can,lam2018xview}.
Representative datasets such as DOTA~\cite{xia2018dota} and DIOR~\cite{li2020object} establish standardized baselines for multi-class object detection, with patch-based processing commonly adopted for training and evaluation.
More recently, SODA-A~\cite{cheng2023towards} extends small-object detection to large aerial scenes, while STAR~\cite{li2025star} introduces scene graph generation on gigapixel satellite imagery, emphasizing the importance of large-scale spatial relationships.
Although some of the above datasets include renewable energy facilities among their diverse classes, these facilities remain peripheral categories within broad object recognition, lacking dedicated evaluation protocols and tailored benchmarks for clean-energy facility mapping in large-size satellite imagery.

\noindent \textbf{Clean-Energy Facility Datasets.}
Existing clean-energy datasets vary considerably in spatial resolution, annotation granularity, and geographic coverage, as summarized in Table~\ref{tab:wind_solar_datasets}.
For wind turbines, localized aerial collections~\cite{kornfein2023closing,hao_mi_2023_7808269} provide bounding-box annotations but remain geographically confined, whereas large-scale resources such as TinyWT~\cite{zhu2024tinywt} and SatlasPretrain~\cite{bastani2023satlaspretrain} offer broader coverage yet rely on point-level localization annotations that lack explicit object extents.
For solar panels, high-resolution datasets such as DukePV~\cite{bradbury2016distributed} and BDAPPV~\cite{kasmi2023crowdsourced} provide detailed annotations from localized aerial imagery, whose imaging characteristics differ from those of orbital satellite observations.
Meanwhile, GloSoFarID~\cite{yang2024glosofarid} expands geographic coverage using 10\,m Sentinel-2 imagery with partially model-generated annotations.
Large-scale mapping products such as ChinaPV~\cite{liu2024vectorized} (30\,m) and World-PV~\cite{zeng2025global} (10\,m) further provide valuable regional and global solar installation inventories for geographic mapping and spatiotemporal analysis.
Despite these advances, systematic benchmarking of fine-grained clean-energy facility recognition on large-size VHR satellite imagery remains limited.
WINSO addresses this gap with 10,962 large-size satellite images and 166,473 expert-verified instances, supporting wind turbine detection and solar panel segmentation within a unified benchmark.

\begin{figure*}[!t]
\centering
\includegraphics[width=\linewidth]{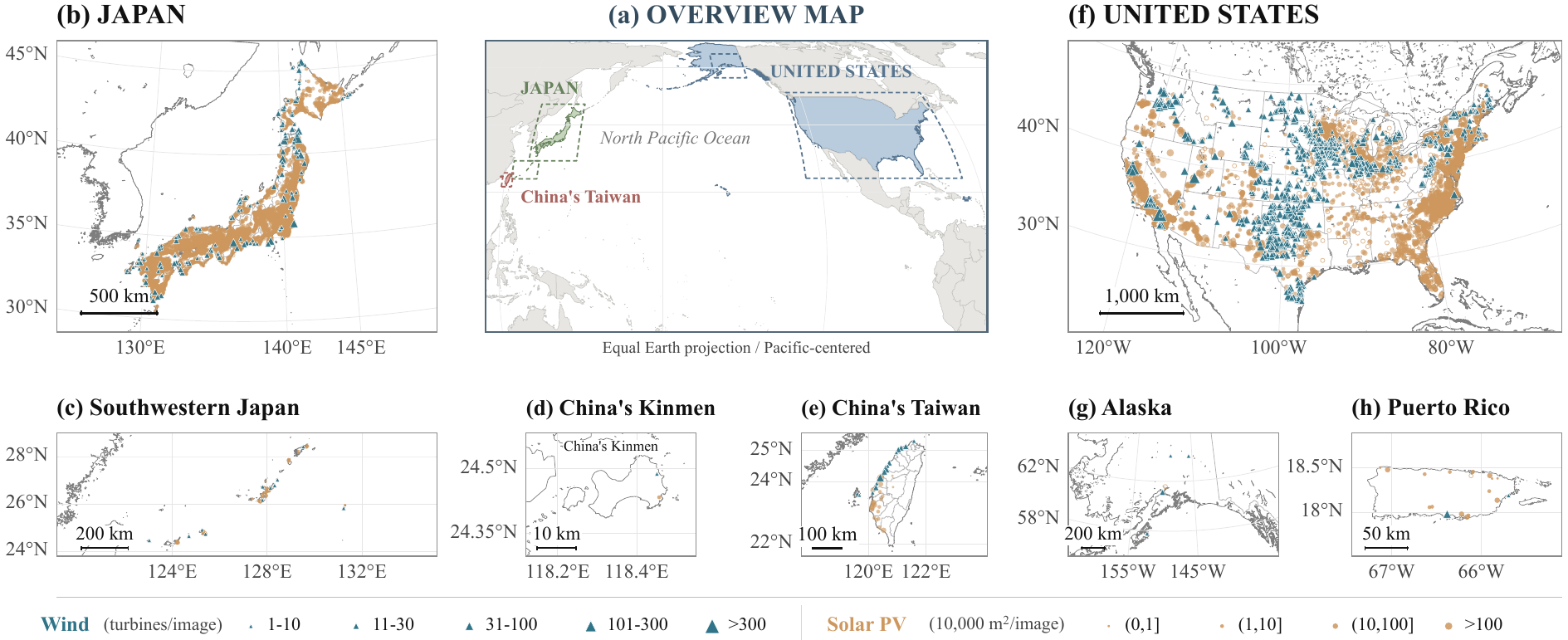} 
\caption{Geographic distribution of the sampled sites in WINSO.}
\label{fig:distribution}
\end{figure*}

\noindent \textbf{Detection and Segmentation on Large-Size Imagery.}
Recent deep detectors and segmentors~\cite{ren2015faster,carion2020end,xie2021segformer,cheng2022masked} deliver strong recognition baselines on standard inputs.
However, deploying them to large-size satellite imagery exposes a fundamental trade-off between computational budget and spatial fidelity.
Direct whole-image downsampling dramatically reduces inference overhead but severely degrades small-object visibility and solar panel boundary details. 
Conversely, tiled (sliding-window) inference preserves fine-grained local cues but fragments macro-context and incurs redundant computation across overlapping regions. 
To reconcile this tension, specialized UHR methods explore selective region processing~\cite{chen2022towards,zhang2023skimming}, global--local feature exchange~\cite{chen2019collaborative}, and shallow--deep feature integration~\cite{guo2022isdnet}.
In parallel, emerging VLMs such as LocateAnything~\cite{wang2026locateanything} and Rex-Omni~\cite{jiang2026detect} introduce language-conditioned zero-shot localization, yet their efficiency and reliability on gigapixel Earth observation scenes remain insufficiently explored. 
WINSO establishes a standardized benchmark protocol for systematically evaluating gloabl inference, tiled inference, UHR architectures, and VLMs in terms of accuracy, inference latency, and GPU memory consumption.
\section{The WINSO Dataset}
\label{sec:winso_dataset}
Our motivation for developing WINSO is twofold: (i) to fill the void of large-scale benchmarks for clean-energy facilities in large-size VHR satellite imagery; and (ii) to promote two meaningful yet challenging tasks: sparse wind turbine detection and fine-grained photovoltaic segmentation in large-size satellite images. To this end, this section provides a comprehensive overview of the WINSO dataset across four key aspects: data collection, annotation pipeline, statistical properties, and dataset splits.

\subsection{Image Collection}
The WINSO dataset comprises 10,962 VHR satellite images with GSDs of 0.5--0.7\,m, purchased from commercial satellite imagery providers and licensed for redistribution as part of the benchmark. All images cover spatially non-overlapping geographic areas, eliminating geographic overlap between dataset splits. In total, WINSO contains 286.36\,Gpx and covers at least 71.6K\,km$^2$, conservatively estimated using the minimum GSD of 0.5\,m. To ensure geographic diversity, we collect imagery from three regions across the Pacific Rim: the United States, Japan, and China's Taiwan region, as shown in Fig.~\ref{fig:distribution}.

\subsection{Annotation Pipeline}
WINSO provides task-specific annotations for two types of clean-energy facilities:
(i)~\textit{Wind Turbines:} Each turbine is annotated with a tight horizontal bounding box (HBB) enclosing the hub, tower base, and visible blade sweep.
(ii)~\textit{Photovoltaic Panels:} PV installations are annotated with pixel-accurate polygon masks, covering utility-scale solar farms, mountainous and agricultural installations, and commercial/industrial rooftop arrays. Each contiguous solar array is delineated along its visible perimeter, excluding inter-array service roads, inverters, and substations to preserve accurate facility boundaries.

\noindent\textbf{Annotation Workflows.}
To facilitate efficient annotation of large-size imagery while maintaining the confidentiality of commercial satellite data, all annotators used locally deployed instances of the open-source Computer Vision Annotation Tool (CVAT)~\cite{CVAT_2023}. This setup enabled image annotation without transferring proprietary imagery to external services. The annotation procedures were tailored to the image characteristics of each task:
\begin{itemize}[leftmargin=0pt,itemindent=\parindent,labelsep=0.5em,noitemsep]
    \item \textit{Annotation for WT:} WT imagery frequently reaches gigapixel dimensions. To accommodate large-image processing and avoid truncating objects at tile boundaries, images exceeding 10,000$\times$10,000 pixels were temporarily divided into overlapping tiles with a 20\% overlap ratio. The resulting bounding boxes were mapped back to the original image coordinates, and duplicate annotations in overlapping regions were removed using boundary-aware non-maximum suppression (NMS). The original large-size images were retained in the final benchmark.
    \item \textit{Annotation for PV:} PV imagery typically spans several thousand pixels per dimension, allowing polygon annotation directly on the original images. This preserves continuous facility boundaries without introducing tile-stitching artifacts.
\end{itemize}

\begin{figure*}[!t]
  \centering
  \begin{subfigure}[t]{0.245\textwidth}
    \centering
    \includegraphics[width=\linewidth]{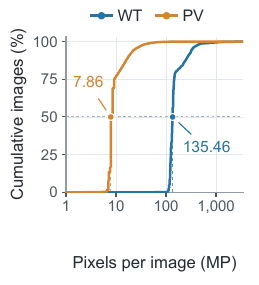}
    \vspace{-6mm}
    \caption{}
    \label{fig:statistics_image_size}
  \end{subfigure}\hfill
  \begin{subfigure}[t]{0.245\textwidth}
    \centering
    \includegraphics[width=\linewidth]{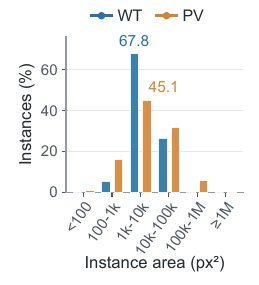}
    \vspace{-6mm}
    \caption{}
    \label{fig:statistics_instance_area}
  \end{subfigure}\hfill
  \begin{subfigure}[t]{0.245\textwidth}
    \centering
    \includegraphics[width=\linewidth]{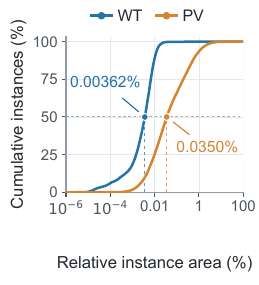}
    \vspace{-6mm}
    \caption{}
    \label{fig:statistics_relative_area}
  \end{subfigure}
  \begin{subfigure}[t]{0.245\textwidth}
    \centering
    \includegraphics[width=\linewidth]{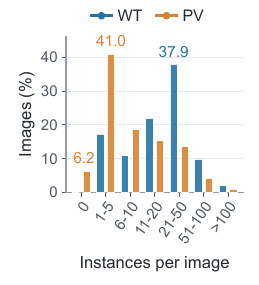}
    \vspace{-6mm}
    \caption{}
    \label{fig:statistics_instance_count}
  \end{subfigure}\hfill
  \vspace{-2mm}
  \caption{Statistics of WINSO for WT and PV.
  From left to right: cumulative distributions of original image pixel counts; instance area distributions; relative instance area distributions; and distributions of instances per image.
  Instance area denotes HBB area for WT and mask pixel area for PV. 
  MP denotes $10^6$ pixels.}
  \label{fig:dataset_statistics}
\end{figure*}

\noindent\textbf{Quality Verification.}
A total of 25 trained annotators were divided into two specialized teams (15 for WT and 10 for PV), completing the annotations over two months. Three remote sensing experts conducted biweekly quality checks and spent an additional month reviewing the entire dataset image by image.
Ambiguous instances were further examined using multi-temporal satellite imagery and historical land-use records before final acceptance.
% Additional details of the annotation pipeline and visual examples from the WINSO are provided in the \textcolor{cvprblue}{Appendix}.

\subsection{Dataset Statistics and Properties}
Compared to existing renewable energy benchmarks, WINSO exhibits significant advantages in spatial resolution, full-scene dimensions, instance diversity, and geographic breadth. 
We characterize these advantages across six core dimensions.

\begin{itemize}[leftmargin=0pt,itemindent=\parindent,labelsep=0.5em,noitemsep]
\item \textbf{\textit{Large Image Sizes.}} 
As shown in Fig.~\ref{fig:statistics_image_size}, WINSO preserves VHR satellite scenes, with over 500 images exceeding 10,000 $\times$ 10,000 pixels and the largest reaching an extreme dimension of 44,032 $\times$ 62,208 pixels. 
Because these images vary substantially in dimension and cover immense spatial extents, conventional downsampling methods with fixed scales are ill-suited to balance global context and local fidelity. 
How to efficiently process such large-size scenes while preventing boundary slicing artifacts and preserving both macro-spatial layouts and fine-grained structural details poses a major challenge for current vision models.

\item \textbf{\textit{Diverse Instance Scales.}} 
As shown in Fig.~\ref{fig:statistics_instance_area}, WINSO encompasses an exceptionally wide spectrum of target scales in terms of absolute pixel area. 
Turbine bounding boxes span three orders of magnitude, whereas photovoltaic mask areas range from 5 to over $4.46\times10^{6}$ pixels, traversing nearly six orders of magnitude. 
The coexistence of microscopic isolated targets alongside massive contiguous structures demands that architectures resolve fine-grained local details of tiny instances while accurately tracing the expansive geometric boundaries of large installations.

\item \textbf{\textit{Tiny Relative Target Footprints.}} 
Despite the substantial absolute pixel counts of certain centralized facilities, their footprints remain vanishingly small in proportion to the gigapixel scenes. 
As shown in Fig.~\ref{fig:statistics_relative_area}, the median ratios of target area to total image area for wind turbines and solar panels are merely 0.0036\% and 0.0350\%, respectively. 
This sharp discrepancy between vast background contexts and minimal relative target coverage means that standard global resizing easily obliterates critical structural cues, necessitating frameworks capable of combining rapid wide-area filtering with fine-grained localization.

\item \textbf{\textit{Variation in Instance Density.}} 
As shown in Fig.~\ref{fig:statistics_instance_count}, the facility count per scene varies substantially across WINSO. 
Centralized renewable bases on expansive plains exhibit dense array deployments, whereas rugged topographies constrain mountain and rural installations to isolated, sparse distributions dominated by extensive backgrounds. 
This significant disparity ensures the dataset encompasses both crowded and sparse scenarios, providing diverse data support to evaluate the robustness of detection and segmentation algorithms under varying density conditions.

\item \textbf{\textit{Diverse Background Types.}} 
WINSO encompasses renewable facilities situated across distinct geographical and terrain backgrounds, including agricultural fields, vegetated plains, arid deserts, mountainous valleys, urban settlements, and water bodies. 
This demands that perception models maintain robust recognition capabilities across varied and complex environmental conditions. 
Furthermore, facilities often visually confound or spatially intertwine with surrounding farmland ridges, building rooftops, and transportation networks, further compounding the difficulty of accurate detection and fine-grained segmentation.

\item \textbf{\textit{Cross-Continental Geographic Coverage.}} 
Sampling sites in WINSO span both sides of the Pacific, covering over 5,000 cities across Asia and North America, with the wind turbine and photovoltaic subsets spanning 807 and 4,454 cities, respectively. 
This extensive geographic spread introduces substantial variations in land-use policies, climatic surface characteristics, and engineering layouts. 
Such widespread regional diversity dramatically enriches visual appearances and environmental contexts, providing an authentic testbed for evaluating the out-of-domain transferability and geographic generalization of vision models.
\end{itemize}

\subsection{Dataset Splits}
To keep the distributions of training and test data consistent, we randomly allocate 1/2 of the original images for training, 1/6 for validation, and 1/3 for testing. 
All images and ground-truth annotations are released for the training and validation sets, while test labels are withheld.
For testing, we are currently deploying an automated evaluation server.
% Detailed split statistics are provided in the \textcolor{cvprblue}{Appendix}.
\section{Experiments}
\label{sec:experiments}
\subsection{Experimental Setup}

\noindent\textbf{Compared Baselines.} 
To comprehensively evaluate performance on our benchmark, we compare task-specific models against VLMs across both tracks:
\begin{itemize}[leftmargin=*,noitemsep,topsep=1pt]
    \item \textit{General Detection Models:} We evaluate standard object detectors across single-stage architectures such as YOLOv5~\cite{jocher2022ultralytics} and YOLOv8~\cite{YOLOv8_2023}, two-stage models like Faster R-CNN~\cite{ren2015faster}, and transformer-based paradigms including DETR~\cite{carion2020end} and RT-DETR~\cite{zhao2024detrs}, alongside UHR detectors such as SparseFormer~\cite{li2024sparseformer}, GigaDET~\cite{chen2022towards}, SPDet~\cite{zhang2023skimming}, and UHR-DETR~\cite{li2026uhr}.
    \item \textit{General Segmentation Models:} For dense prediction, we benchmark general semantic segmentation frameworks including U-Net~\cite{ronneberger2015u}, SegFormer~\cite{xie2021segformer}, DeepLabv3+~\cite{chen2018encoder}, and Mask2Former~\cite{cheng2022masked}, alongside dedicated UHR architectures like GLNet~\cite{chen2019collaborative}, ISDNet~\cite{guo2022isdnet}, and RUE~\cite{qin2025boosting}.
    \item \textit{Vision-Language Models:} To assess the empirical effectiveness of VLMs on this specialized domain, we benchmark LocateAnything-3B~\cite{wang2026locateanything}, Rex-Omni-3B~\cite{jiang2026detect}, Grounding DINO-T~\cite{liu2024grounding}, and Qwen3-VL-4B~\cite{bai2025qwen3}. For detection, these models directly output bounding boxes. For segmentation, their predicted boxes are fed to SAM-2.1~\cite{ravi2025sam} as prompts to generate masks.
\end{itemize}

\noindent\textbf{Evaluation Metrics.} 
For WT detection, we adopt standard COCO metrics, reporting overall $\text{AP}$ across Intersection-over-Union (IoU) thresholds from 0.50 to 0.95, along with $\text{AP}_{50}$, $\text{AP}_{75}$, and scale-stratified metrics $\text{AP}_s$, $\text{AP}_m$, and $\text{AP}_l$, corresponding to bounding box areas within $[0, 32^2]$, $[32^2, 96^2]$, and $[96^2, +\infty)$ pixels, respectively. 
For PV segmentation, we report foreground IoU, mean IoU (mIoU) over the foreground and background classes, and foreground F1-score.
Operational efficiency is benchmarked using per-image inference latency in seconds and peak GPU memory footprint in gigabytes.

\noindent\textbf{Implementation Details.} 
All baselines are implemented via standard open-source libraries or their official releases: 
Faster R-CNN and DETR are built on MMDetection~\cite{mmdetection},
YOLO variants on Ultralytics~\cite{Jocher_Ultralytics}, 
SegFormer, ISDNet, and RUE on MMSegmentation~\cite{mmseg2020}, 
and Mask2Former on Detectron2~\cite{wu2019detectron2}. 
General detectors are initialized from COCO-pretrained checkpoints, while segmentation backbones, including ResNet50~\cite{he2016deep} and MiT~\cite{xie2021segformer} variants, are pretrained on ImageNet-1K~\cite{deng2009imagenet}. 
For WT detection, general detectors are evaluated under two inference protocols: direct global resizing to 1024$\times$1024 and sliding-window cropping with 1024$\times$1024 patches at 20\% overlap. Their training follows the standard $1\times$ schedule under MMDetection or 100 epochs under Ultralytics, while RT-DETR and dedicated UHR models maintain their default configurations. 
For PV segmentation, standard models train for 20 epochs on 512$\times$512 crops with a batch size of 8, whereas UHR methods train for 40k iterations using 2000$\times$2000 patches with a batch size of 1. 
For VLMs, inputs are processed as 2048$\times$2048 crops for inference. 
All models are trained on two NVIDIA RTX 3090 GPUs with 24GB VRAM each, while inference latency and memory footprints are benchmarked on a single GPU. 
% Detailed implementation settings are provided in the \textcolor{cvprblue}{Appendix}.

\begin{table*}[!t]
\centering
\caption{Detection performance on the WT benchmark.}
\label{tab:detection_comparison}
\scriptsize
\resizebox{\textwidth}{!}{%
  \begin{tabular}{@{}l|c|*{6}{c|}c|c@{}}
    \toprule
    Method & Backbone
    & $\mathrm{AP}$ $\uparrow$
    & $\mathrm{AP}_{50}$ $\uparrow$
    & $\mathrm{AP}_{75}$ $\uparrow$
    & $\mathrm{AP}_{\mathrm{s}}$ $\uparrow$
    & $\mathrm{AP}_{\mathrm{m}}$ $\uparrow$
    & $\mathrm{AP}_{\mathrm{l}}$ $\uparrow$
    & Latency (s) $\downarrow$
    & Mem. (GB) $\downarrow$ \\
    \midrule
    \multicolumn{9}{@{}l}{\textbf{\textit{Local Inference}}} \\
      YOLOv5-m~\cite{jocher2022ultralytics}  & CSPDarknet   & \underline{40.83} & \underline{78.31} & \textbf{37.48} & \textbf{6.74} & \textbf{42.39} & \underline{51.83} & 14.99 & 0.65 \\
      YOLOv8-m~\cite{YOLOv8_2023}            & Modified CSP & \textbf{41.08} & \textbf{78.53} & \underline{37.41} & \underline{6.61} & \underline{42.17} & \textbf{52.67} & 15.61 & 0.68 \\
      Faster R-CNN~\cite{ren2015faster}      & ResNet50-FPN & 36.39 & 74.96 & 29.91 & 3.79 & 37.75 & 47.19 & 11.59 & 1.74 \\
      DETR~\cite{carion2020end}              & ResNet50     & 22.37 & 60.10 & 11.35 & 0.26 & 21.49 & 38.27 & 16.40 & 9.83 \\
      RT-DETR~\cite{zhao2024detrs}           & ResNet50     & 37.34 & 72.95 & 33.34 & 4.18 & 37.31 & 49.46 & 9.24 & 12.97 \\
    \midrule
    \multicolumn{9}{@{}l}{\textbf{\textit{Global Inference}}} \\
      YOLOv5-m~\cite{jocher2022ultralytics}  & CSPDarknet   & 9.27 & 33.18 & 2.59 & 0.00 & 6.90 & 18.28 & \underline{0.65} & \textbf{0.57} \\
      YOLOv8-m~\cite{YOLOv8_2023}            & Modified CSP & 9.51 & 33.17 & 2.74 & 0.00 & 6.91 & 18.70 & \textbf{0.64} & \underline{0.60} \\
      Faster R-CNN~\cite{ren2015faster}      & ResNet50-FPN & 0.04  & 0.18 & 0.01 & 0.00 & 0.00 & 0.08  & \underline{0.65} & 1.22 \\
      DETR~\cite{carion2020end}              & ResNet50     & 0.05  & 0.28 & 0.00 & 0.00 & 0.48 & 0.14  & 0.67 & 1.02 \\
      RT-DETR~\cite{zhao2024detrs}           & ResNet50     & 1.37  & 6.02 & 0.06 & 0.00 & 0.90 & 3.22  & 0.66 & 1.25 \\
    \midrule
    \multicolumn{9}{@{}l}{\textbf{\textit{UHR Methods}}} \\
      SparseFormer~\cite{li2024sparseformer} & SparseNet  & 23.13 & 60.30 & 14.09 & 1.88 & 22.34 & 34.84 & 47.75 & 1.46 \\
      GigaDET~\cite{chen2022towards}        & ResNet50   & 26.71 & 56.56 & 21.45 & 0.32 & 25.25 & 40.93 & 0.80 & 2.65 \\
      SPDet~\cite{zhang2023skimming}        & CSPDarknet & 35.79 & 74.09 & 29.75 & 2.73 & 36.72 & 47.95 & 6.42 & 3.41 \\
      UHR-DETR~\cite{li2026uhr}             & ResNet50   & 22.84 & 49.96 & 17.15 & 0.54 & 21.60 & 35.36 & 0.74 & 10.87 \\
    \midrule
    \multicolumn{9}{@{}l}{\textbf{\textit{Vision-Language Models}}} \\
      Grounding DINO-T~\cite{liu2024grounding}              & Swin-T & 0.05 & 0.17 & 0.02 & 0.00 & 0.64 & 0.04 & 23.30  & 1.87 \\
      Rex-Omni-3B~\cite{jiang2026detect}                    & -- & 0.90 & 3.72 & 0.09 & 0.39 & 1.59 & 0.38 & 126.90 & 8.89 \\
      Qwen3-VL-4B~\cite{bai2025qwen3}                       & -- & 0.23 & 0.81 & 0.09 & 0.00 & 0.45 & 0.17 & 371.61 & 11.00 \\
      LocateAnything-3B~\cite{wang2026locateanything}       & -- & 0.50 & 1.39 & 0.29 & 0.64 & 1.36 & 0.26 & 195.58 & 14.90 \\
    \bottomrule
  \end{tabular}%
}
\end{table*}

\begin{figure*}[!t]
\centering
\includegraphics[width=\linewidth]{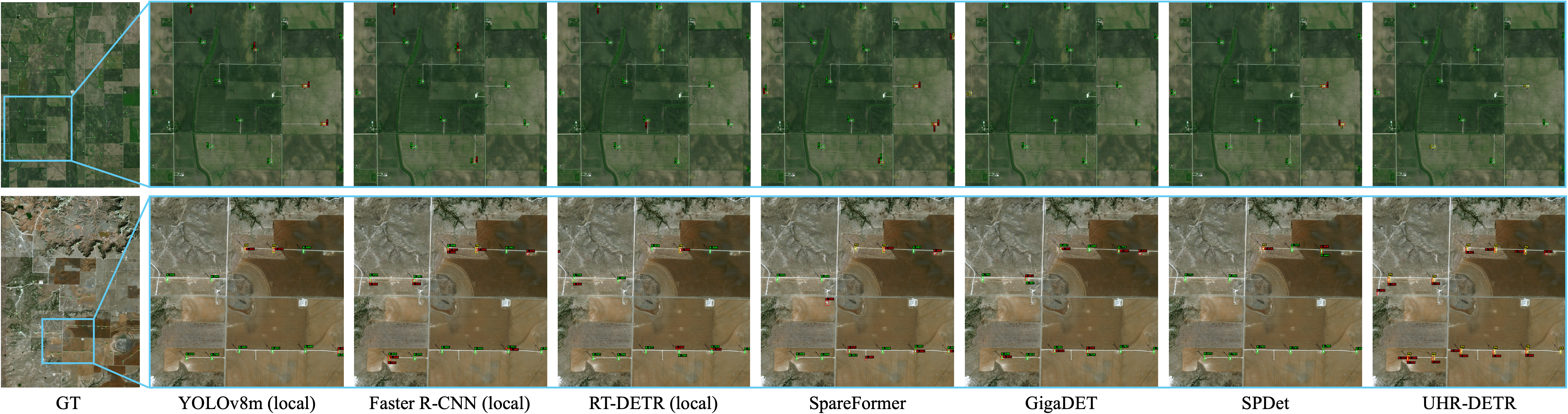} 
\caption{Qualitative results of WT detection.
Green, red and yellow boxes represent correct, incorrect and missed detection, respectively.}
\label{fig:visual_wind}
\end{figure*}

\subsection{Results on WT Detection}
Tab.~\ref{tab:detection_comparison} summarizes the performance of various detection paradigms on the WT benchmark.
Across standard detectors, local sliding-window inference consistently outperforms whole-image resizing, with YOLOv8-m achieving the highest accuracy of 41.08\% $\text{AP}$ and 78.53\% $\text{AP}_{50}$.
However, this accuracy comes at the cost of substantial latency, requiring 15.61\,s per image.
Conversely, direct global downsampling reduces latency to approximately 0.65s but significantly degrades detection accuracy.
Even under local inference, the highest $\text{AP}_s$ is only 6.74\%, indicating that small-target detection remains a notable bottleneck.
Several UHR methods offer competitive accuracy--efficiency trade-offs: SPDet attains 35.79\% $\text{AP}$ at 6.42\,s, representing a 2.43$\times$ speedup over local YOLOv8-m, while GigaDET and UHR-DETR achieve sub-second inference at 0.80\,s and 0.74\,s, respectively.
Nonetheless, these UHR architectures generally struggle with small turbines, consistently scoring below 3\% in $\text{AP}_s$.
In particular, GigaDET and UHR-DETR selectively process high-priority regions, which may miss turbines dispersed across extensive wind-farm scenes under limited computational budgets.
One promising direction is candidate-point-guided local refinement, focusing high-resolution processing on potential turbine locations.
The evaluated VLMs also exhibit limited accuracy with high inference times.
Qualitative comparisons in Fig.~\ref{fig:visual_wind} further illustrate the missed detections of small turbines by efficient UHR models.
Overall, these results highlight the importance of more effective region selection strategies for efficient detection in large-size remote sensing imagery.

\subsection{Results on PV Segmentation}
\begin{figure*}[!t]
\centering
\includegraphics[width=\linewidth]{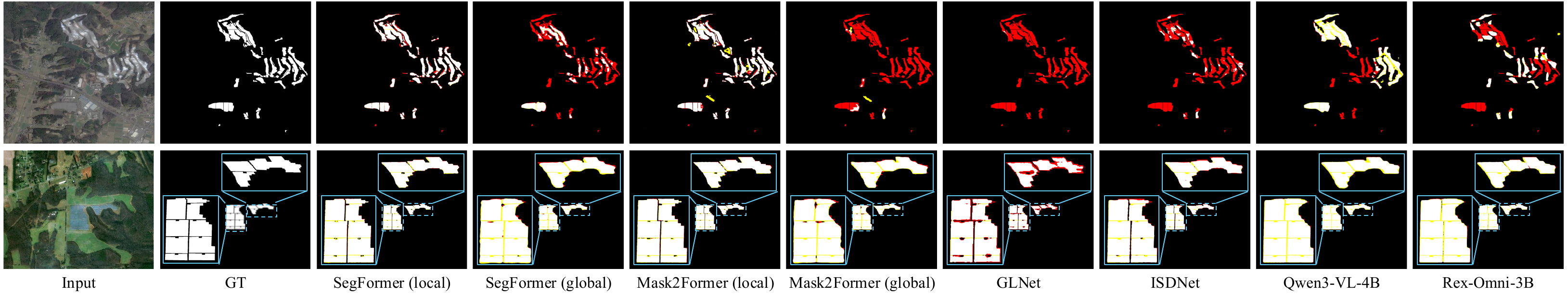} 
\caption{Qualitative results of PV segmentation. White, yellow, and red regions indicate correctly predicted, false positives, and false negatives, respectively.}
\label{fig:visual_solar}
% \vspace{-5mm}
\end{figure*}

Tab.~\ref{tab:segment_comparison} summarizes the PV segmentation results across different paradigms. 
Under local inference, SegFormer and Mask2Former achieve the highest foreground IoU of 86.92\% and 86.73\%, respectively. 
Global inference also achieves strong aggregate performance, with UNet attaining 84.60\% IoU at a latency of only 0.04\,s per image. 
Despite these strong aggregate scores, qualitative comparisons reveal failures on small or fragmented installations.
As shown in Fig.~\ref{fig:visual_solar}, global baselines generally recover the primary extent of large contiguous arrays with minor boundary and inter-array gap errors, yet miss fragmented installations in complex terrain, where local inference better preserves fine spatial details. 
Dedicated UHR models such as GLNet and ISDNet also exhibit noticeable false negatives in these fragmented scenes, indicating that their high-resolution mechanisms do not fully resolve the challenge of scattered small targets. 
Meanwhile, the evaluated VLM pipelines tend to produce over-segmented masks and imprecise boundaries alongside substantial computational overhead. 
Together, these observations suggest exploring adaptive global--local inference for PV segmentation, drawing inspiration from selective region processing in UHR detection. Such an approach could combine efficient global processing of large contiguous arrays with targeted high-resolution refinement of small or fragmented installations.

\begin{table}[!t]
\centering
\caption{Segmentation performance on the PV benchmark.}
\label{tab:segment_comparison}
\resizebox{\linewidth}{!}{
\begin{tabular}{lcccccc}
  \toprule
  Method
  & Backbone 
  & \makecell{$\mathrm{IoU}$ $\uparrow$}
  & \makecell{$\mathrm{mIoU}$ $\uparrow$}
  & \makecell{$\mathrm{F1}$ $\uparrow$}
  & \makecell{Latency\\(s) $\downarrow$} 
  & \makecell{Mem.\\(GB) $\downarrow$} \\
  \midrule
  \textit{\textbf{Local Inference}} \\
  UNet~\cite{ronneberger2015u}       & ResNet50 & 83.54 & 91.43 & 91.03 & 0.77 & 1.62 \\
  SegFormer~\cite{xie2021segformer}  & MiT-B1   & \textbf{86.92} & \textbf{93.20} & \textbf{93.00} & 1.34 & 5.54 \\
  Deeplabv3+~\cite{chen2018encoder}  & ResNet50 & 83.96 & 91.66 & 91.28 & 0.55 & 0.98 \\
  Mask2Former~\cite{cheng2022masked} & ResNet50 & \underline{86.73} & \underline{93.10} & \underline{92.90} & 0.99 & 1.83 \\
  \midrule
  \textit{\textbf{Global Inference}} \\
  UNet~\cite{ronneberger2015u}       & ResNet50 & 84.60 & 92.00 & 91.66 & \textbf{0.04} & \underline{0.29} \\
  SegFormer~\cite{xie2021segformer}  & MiT-B1   & 82.67 & 91.00 & 90.52 & 0.22 & 2.72 \\
  Deeplabv3+~\cite{chen2018encoder}  & ResNet50 & 74.22 & 86.61 & 85.21 & \underline{0.06} & \textbf{0.27} \\
  Mask2Former~\cite{cheng2022masked} & ResNet50 & 83.87 & 91.62 & 91.23 & 0.09 & 0.43 \\
  \midrule
  \textit{\textbf{UHR Methods}} \\
  GLNet~\cite{chen2019collaborative} & ResNet50 & 77.72 & 88.43 & 87.46 & 6.74 & 2.33 \\
  ISDNet~\cite{guo2022isdnet}        & -        & 79.77 & 89.50 & 88.75 & 6.74 & 4.64 \\
  RUE~\cite{qin2025boosting}         & -        & 60.96 & 79.65 & 75.74 & 0.60 & 4.14 \\
  \midrule
  \textit{\textbf{Vision-Language Models}} \\
  Grounding DINO-T~\cite{liu2024grounding}        & Swin-T & 29.24 & 62.49 & 45.25 & 0.44  & 3.03 \\
  Rex-Omni-3B~\cite{jiang2026detect}              & --     & 65.66 & 82.13 & 79.27 & 8.61  & 10.62 \\
  Qwen3-VL-4B~\cite{bai2025qwen3}                 & --     & 68.02 & 83.29 & 80.97 & 27.15 & 12.91 \\
  LocateAnything-3B~\cite{wang2026locateanything} & --     & 44.07 & 70.10 & 61.18 & 43.94 & 14.89 \\
  \bottomrule
  \end{tabular}%
  }
\end{table}

\begin{table}[!t]
\centering
\caption{Cross-dataset generalization performance.}
\label{tab:cross_dataset}
\resizebox{\linewidth}{!}{
\begin{tabular}{llcccc}
  \toprule
  $\mathrm{Test\ on}$
  & $\mathrm{Train\ on}$
  & \makecell{$\mathrm{P}$ $\uparrow$}
  & \makecell{$\mathrm{R}$ $\uparrow$}
  & \makecell{$\mathrm{F1}$ $\uparrow$} 
  & \makecell{$\mathrm{AP}_{50}$ $\uparrow$} \\
  \midrule
  \multirow{3}{*}{STAR~\cite{li2025star}}             
    & DIOR~\cite{li2020object} & 63.47 & 95.56 & 76.27 & 94.07 \\
    & \multirow{2}{*}{WINSO \textbf{(Ours)}}    & 85.51 & 98.33 & 91.47 & 95.42 \\
    & & \textcolor{darkgreen}{$\uparrow$ 22.04} & \textcolor{darkgreen}{$\uparrow$ 2.77} & \textcolor{darkgreen}{$\uparrow$ 15.20} & \textcolor{darkgreen}{$\uparrow$ 1.35} \\
  \toprule
  $\mathrm{Test\ on}$
  & $\mathrm{Train\ on}$
  & \multicolumn{2}{c}{$\mathrm{mIoU}$ $\uparrow$} 
  & \multicolumn{2}{c}{$\mathrm{F1}$ $\uparrow$} \\
  \midrule
  \multirow{2}{*}{PVP-Dataset~\cite{wang2023pvnet}}   
    & AIR-PV~\cite{yan2023air} 
    & \multicolumn{2}{c}{45.87} & \multicolumn{2}{c}{25.80} \\
    & WINSO \textbf{(Ours)} 
    & \multicolumn{2}{c}{\textbf{68.50} \textcolor{darkgreen}{$\uparrow$ 22.63}} & \multicolumn{2}{c}{\textbf{74.23} \textcolor{darkgreen}{$\uparrow$ 48.43}} \\
  \bottomrule
\end{tabular}%
}
\end{table}

\subsection{Cross-Dataset Generalization}
Tab.~\ref{tab:cross_dataset} evaluates the cross-dataset generalization of WINSO using Faster R-CNN for wind turbine detection and U-Net for solar panel segmentation. On the STAR test set, Faster R-CNN trained on WINSO outperforms its DIOR-trained counterpart across all metrics, achieving gains of 22.04\% in precision and 15.20\% in F1-score. Similarly, on PVP-Dataset, U-Net trained on WINSO surpasses the AIR-PV-trained baseline by 22.63\% in mIoU and 48.43\% in F1-score. These results demonstrate the strong transferability of WINSO-trained models to external datasets, highlighting its value for robust clean-energy facility mapping. 
% Further implementation details are provided in the \textcolor{cvprblue}{Appendix}.
\section{Discussion and Outlook}
\label{sec:discussion}
\noindent\textit{\textbf{Benchmark Value.}} 
WINSO establishes a standardized benchmark grounded in large-size satellite observations rather than pre-partitioned tiles.
Preserving wide-area context exposes the coupled challenges of slender turbine structures, complex solar array geometries, and heterogeneous natural backgrounds.
This setting provides an authentic foundation to jointly evaluate recognition fidelity and computational efficiency for renewable energy facility mapping.

\noindent\textit{\textbf{Application Value.}} 
The detailed geometric boundaries and spatial distributions in WINSO provide essential physical parameters for downstream planning and governance. 
Accurately resolving wind turbine and solar panel coordinates directly informs infrastructure siting, from transmission corridors and energy storage to power-intensive datacenters. 
More broadly, these high-fidelity spatial inventories support diverse domains, including energy yield estimation~\cite{song2026coal}, power grid dispatch~\cite{hu2026advancing}, land-use compliance auditing~\cite{hu2025quantifying}, and regional carbon accounting~\cite{wang2025increased}.

\noindent\textit{\textbf{Methodological Outlook.}} 
Given the extreme spatial sparsity of renewable facilities across large-format scenes, uniform computation introduces substantial redundancy. 
A promising research frontier lies in content-adaptive architectures that dynamically route dense computations exclusively toward candidate regions. 
Advancing such sparse perception paradigms will better balance recognition fidelity against full-scene throughput, fostering efficient models for wide-area Earth observation.

\section{Conclusion}
\label{sec:conclusion}
In this paper, we present WINSO, a large-scale benchmark for renewable energy facility perception in large-size VHR satellite imagery, supporting both wind turbine detection and solar panel segmentation.
The dataset comprises 10,962 images and 166,473 expert-verified instances spanning over 5,000 cities across Asia and North America.
The large image sizes, substantial sample volume, wide geographic coverage, and extensive diversity in object scale and background type make WINSO a valuable benchmark to advance wind turbine detection and solar array segmentation in satellite imagery.
Leveraging this benchmark, we systematically evaluate general detection models, dedicated UHR architectures, and VLMs across diverse inference regimes, uncovering critical trade-offs between recognition fidelity and computational throughput.
Future work will expand WINSO across broader geographic territories and explore content-adaptive sparse architectures to facilitate clean-energy monitoring on a global scale.
% \newpage
{
    \small
    \bibliographystyle{ieeenat_fullname}
    \bibliography{main}
}

\end{document}